\documentclass[letterpaper]{article}
\usepackage{iccc}

\usepackage{times}
\usepackage{helvet}
\usepackage{courier}
\usepackage{graphicx}
\usepackage{amsmath}
\usepackage{amsfonts}
\usepackage{amssymb}
\usepackage{url}
\usepackage{hyperref}

\newcommand{\argmin}[1]{\underset{#1}{\operatorname{arg}\,\operatorname{min}}\;}
\newcommand{\bv}[1]{\mathbf{#1}}

\title{Semantic Style Transfer and Turning Two-Bit Doodles into Fine Artwork}
\author{
    Alex J. Champandard\\
    nucl.ai Research Laboratory\\
    alexjc@nucl.ai
  \And
    nucl.ai Conference 2016\thanks{This research was funded out of the marketing budget.}\\
    Artificial Intelligence in Creative Industries\\
    July 18-20, Vienna/Austria.\\
    \url{http://events.nucl.ai/}
}
\begin{document} 
\maketitle
\begin{abstract}
\begin{quote}
Convolutional neural networks (CNNs) have proven highly effective at image synthesis and style transfer. For most users, however, using them as tools can be a challenging task due to their unpredictable behavior that goes against common intuitions. This paper introduces a novel concept to augment such generative architectures with semantic annotations, either by manually authoring pixel labels or using existing solutions for semantic segmentation. The result is a content-aware generative algorithm that offers meaningful control over the outcome. Thus, we increase the quality of images generated by avoiding common glitches, make the results look significantly more plausible, and extend the functional range of these algorithms---whether for portraits or landscapes, etc.  Applications include semantic style transfer and turning doodles with few colors into masterful paintings!
\end{quote}
\end{abstract}

\section{Introduction}

Image processing algorithms have improved dramatically thanks to CNNs trained on image classification problems to extract underlying patterns from large datasets~\cite{simonyan14}. As a result, deep convolution layers in these networks provide a more expressive feature space compared to raw pixel layers, which proves useful not only for classification but also generation~\cite{mahendran14}.  For transferring style between two images in particular, results are astonishing---especially with painterly, sketch or abstract styles~\cite{gatys15}.

\begin{figure}[t]
\includegraphics[scale=1.1]{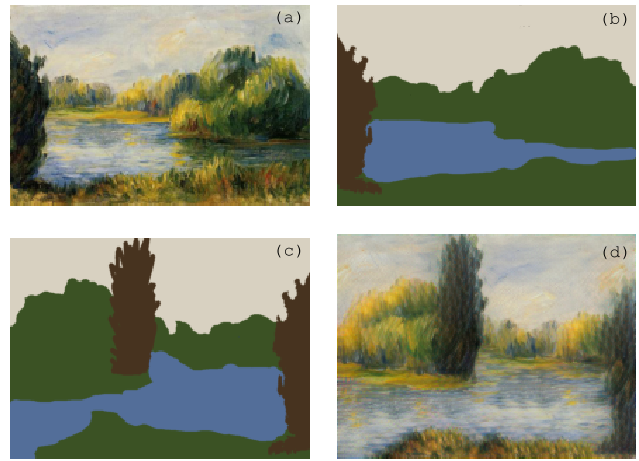}
\caption{Synthesizing paintings with deep neural networks via analogy. (a) Original painting by Renoir, (b) semantic annotations, (c) desired layout, (d) generated output.}
\label{fig:analogy}
\end{figure}

However, to achieve good results using neural style transfer in practice today, users must pay particular attention to the composition and/or style image selection, or risk seeing unpredictable and incorrect patterns. For portraits, facial features can be ruined by incursions of background colors or clothing texture, and for landscapes pieces of vegetation may be found in the sky or other incoherent places. There's certainly a place for this kind of glitch art, but many users become discouraged not being able to get results they want.


Through our social media bot that first provided these algorithms as a service~\cite{champandard15}, we observe that users have clear expectations how style transfer should occur: most often this matches semantic labels, e.g. hair style and skin tones should transfer respectively regardless of color. Unfortunately, while CNNs routinely extract semantic information during classification, such information is poorly exploited by generative algorithms---as evidenced by frequent glitches.

\begin{figure*}[!t]
\includegraphics[scale=1.03]{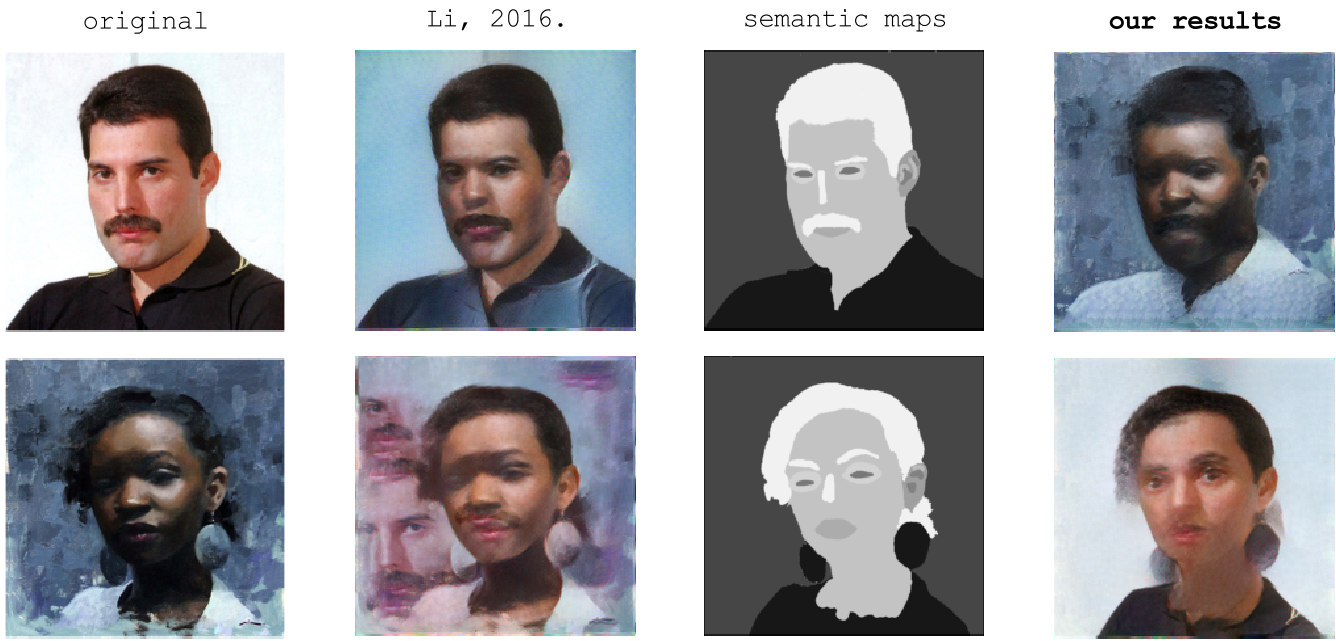}
\caption{Comparison and breakdown of synthesized portraits, chosen because of extreme color and feature mismatches. Parameters were adjusted to make the style transfer most faithful while reducing artifacts such as patch repetition or odd blends---which proved challenging for the second column, but more straightforward in the last column thanks to semantic annotations. The top row shows transfer of painted style onto a photo (easier), and the bottom turning the painting into a photo (harder); see area around the nose and mouth for failures. [Original painting by Mia Bergeron.]}
\label{fig:portraits}
\end{figure*}

We attribute these problems to two underlying causes:

\begin{enumerate}
\item While CNNs used for classification can be re-purposed to extract style features (e.g. textures, grain, strokes), they were not architected or trained for correct synthesis.
\item Higher-level layers contain the most meaningful information, but this is not exploited by the lower-level layers used in generative architectures: only error back-propagation indirectly connects layers from top to bottom.
\end{enumerate}

To remedy this, we introduce an architecture that bridges the gap between generative algorithms and pixel labeling neural networks.  The architecture commonly used for image synthesis~\cite{simonyan14} is augmented with semantic information that can be used during generation. Then we explain how existing algorithms can be adapted to include such annotations, and finally we showcase some applications in style transfer as well as image synthesis by analogy (e.g. Figure \ref{fig:analogy}).

\section{Related Work}

\begin{figure*}[!t]
\centering
\includegraphics[scale=1.1]{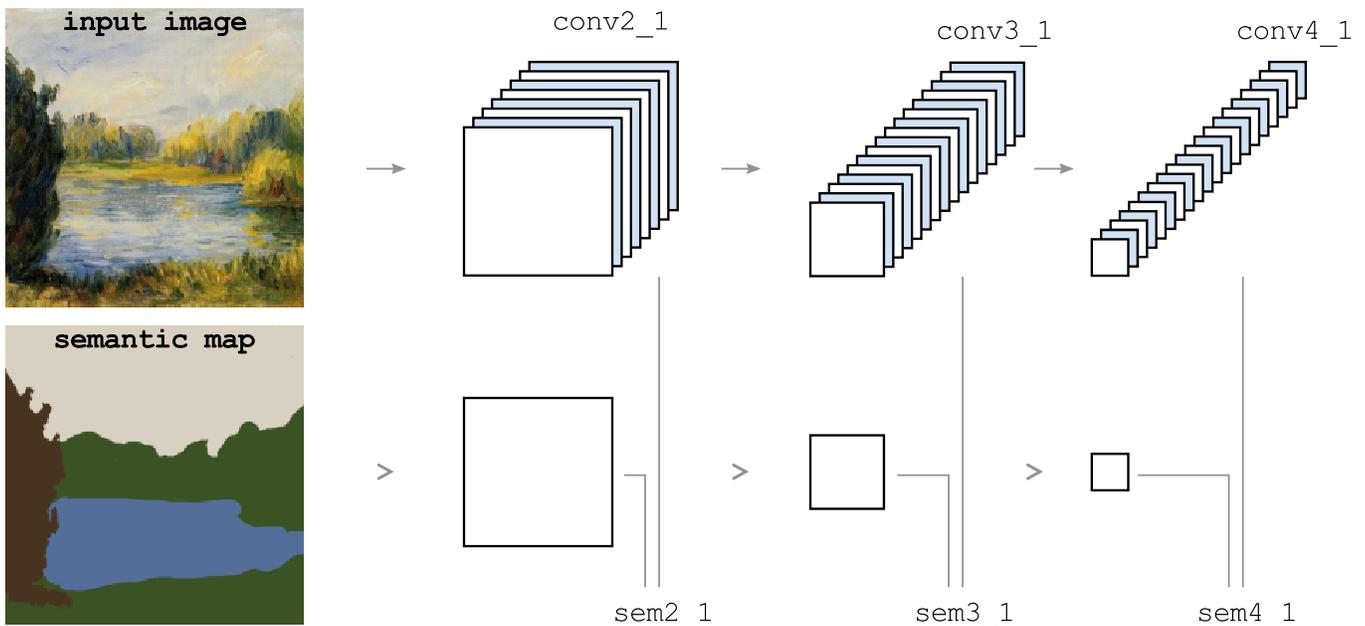}
\caption{Our augmented CNN that uses regular filters of \texttt{N} channels (top), concatenated with a semantic map of \texttt{M=1} channel (bottom) either output from another network capable of labeling pixels or as manual annotations.}
\label{fig:network}
\end{figure*}

The \textbf{image analogy} algorithm \cite{hertzmann01} is able to transfer artistic style using pixel features and their local neighborhoods. While more recent algorithms using deep neural networks generate better quality results from a stylistic perspective, this technique allows users to synthesize new images based on simple annotations. As for recent work on style transfer, it can be split into two categories: specialized algorithms or more general neural approaches.

The first neural network approach to style transfer is \textbf{gram-based}~\cite{gatys15}, using so-called ``Gram Matrices'' to represent global statistics about the image based on output from convolution layers.  These statistics are computed by taking the inner product of intermediate activations---a tensor operation that results in a \(N \times N\) matrix for each layer of \(N\) channels. During this operation, all local information about pixels is lost, and only correlations between the different channel activations remain.  When glitches occur, it's most often due to these global statistics being imposed onto the target image regardless of its own statistical distribution, and without any understanding of local pixel context.

A more recent alternative involves a \textbf{patch-based} approach~\cite{li16}, which also operates on the output of convolution layers.  For layers of \(N\) channels, neural patches of \(3 \times 3\) are matched between the style and content image using a nearest neighbor calculation.  Operating on patches in such a way gives the algorithm local understanding of the patterns in the image, which overall improves the precision of the style transfer since fewer errors are introduced by globally enforcing statistical distributions.

Both gram- and patch-based approaches struggle to provide reliable user controls to help address glitches. The primary parameter exposed is a weighting factor between style and content; adjusting this results in either an abstract-styled mashup that mostly ignores the input content image, or the content appears clearly but its texture looks washed out (see Figure \ref{fig:portraits}, second column).  Finding a compromise where content is replicated precisely and the style is faithful remains a challenge---in particular because the algorithm lacks semantic understanding of the input.

Thankfully, recent CNN architectures are capable of providing such semantic context, typically by performing pixel \textbf{labeling and segmentation} \cite{thoma16}.  These models rely primarily on convolutional layers to extract high-level patterns, then use deconvolution to label the individual pixels. However, such insights are not yet used for synthesis---despite benefits shown by non-neural approaches.

The state-of-the-art \textbf{specialized} approaches to style transfer exploit semantic information to great effect, performing color transfer on photo portraits using specifically crafted image segmentation~\cite{yang15}.  In particular, facial features are extracted to create masks for the image, then masked segments are processed independently and colors can be transferred between each corresponding part (e.g. background, clothes, mouth, eyes, etc.) Thanks to the additional semantic information, even simpler histogram matching algorithms may be used to transfer colors successfully.

\section{Model}

Our contribution builds on a patch-based approach~\cite{li16} to style transfer, using optimization to minimize content reconstruction error $E_c$ (weighted by $\alpha$) and style remapping error $E_s$ (weight $\beta$).  See \cite{gatys15} for details about $E_c$.

\begin{equation}
E = \alpha E_c + \beta E_s
\end{equation}

First we introduce an augmented CNN (Figure~\ref{fig:network}) that incorporates semantic information, then we define the input semantic map and its representation, and finally show how the algorithm is able to exploit this additional information.

\subsection{Architecture}

The most commonly used CNNs for image synthesis is VGG~\shortcite{simonyan14}, which combines pooling and convolution layers $l$ with $3 \times 3$ filters (e.g. the first layer after third pool is named \texttt{conv4\_1}). Intermediate post-activation results are labeled $\bv{x}^l$ and consist of $N$ channels, which capture patterns from the images for each region of the image: grain, colors, texture, strokes, etc. Other architectures tend to skip pixels regularly, compress data, or optimized for classification---resulting in low-quality synthesis~\cite{nikulin16}.

Our augmented network concatenates additional semantic channels $\bv{m}^l$ of size $M$ at the same resolution, computed by down-sampling a static semantic map specified as input. The result is a new output with $N+M$ channels, denoted $\bv{s}^l$ and labeled accordingly for each layer (e.g. \texttt{sem4\_1}).

Before concatenation, the semantic channels are weighted by parameter $\gamma$ to provide an additional user control point:

\begin{equation}
\bv{s}^l = \bv{x}^l \Vert \gamma \bv{m}^l
\label{eq:catn}
\end{equation}

For style images, the activations for the input image and its semantic map are concatenated together as $\bv{s}^l_s$. For the output image, the current activations $\bv{x}^l$ and the input content's semantic map are concatenated as $\bv{s}^l$. Note that the semantic part of this vector is, therefore, static during the optimization process (implemented using L-BFGS).

  This architecture allows specifying manually authored semantic maps, which proves to be a very convenient tool for user control---addressing the unpredictability of current generative algorithms. It also lets us transparently integrate recent pixel labeling CNNs~\cite{thoma16}, and leverage any advances in this field to apply them to image synthesis.

\subsection{Representation}

The input semantic map can contain an arbitrary number of channels $M$.  Whether doing image synthesis or style transfer, there are only two requirements:

\begin{itemize}
\item Each image has its own semantic map of the same aspect ratio, though it can be lower resolution (e.g. 4x smaller) since it'll be downsampled anyway.
\item The semantic maps may use an arbitrary number of channels and representation, as long as they are consistent for the current style and content (so $M$ must be the same).
\end{itemize}

Common representations include single greyscale channels or RGB+A colors---both of which are very easy to author. The semantic map can also be a collection of layer mask per label as output by existing CNNs, or even some kind of ``semantic embedding'' that compactly describes image pixels (i.e. the representation for hair, beards, and eyebrows in portraits would be in close proximity).

\subsection{Algorithm}

Patches of $k \times k$ are extracted from the semantic layers and denoted by the function $\Psi$, respectively $\Psi(\bv{s}_s^l)$ for the input style patches and $\Psi(\bv{s}^l)$ for the current image patches.  For any patch $i$ in the current image and layer $l$, its nearest neighbor $\operatorname{NN}(i)$ is computed using normalized cross-correlation---taking into account weighted semantic map:

\begin{equation}
\operatorname{NN}(i) := \argmin{j}\frac{\Psi_i(\bv{s})\cdot\Psi_j(\bv{s}_{s})}{|\Psi_i(\bv{s})|\cdot|\Psi_j(\bv{s}_{s})|}
\label{eq:ncorr}
\end{equation}

The style error $E_{s}$ between all the patches $i$ of layer $l$ in the current image to the closest style patch is defined as the sum of the Euclidean distances:

\begin{equation}
E_{s}(\bv{s}, \bv{s}_{s}) = \sum_{i}||\Psi_i(\bv{s}) - \Psi_{\operatorname{NN}(i)}(\bv{s}_{s})||^{2}
\label{eq:mrf}
\end{equation}

Note that the information from the semantic map in $\bv{m}^l$ is used to compute the best matching patches and contributes to the loss value, but is not part of the derivative of the loss relative to the current pixels; only the differences in activation $\bv{x^l}$ compared to the style patches cause an adjustment of the image itself via the L-BFGS algorithm.

By using an augmented CNN that's compatible with the original, existing patch-based implementations can use the additional semantic information without changes. If the semantic map and $\bv{m^l}$ is zero, the original algorithm~\shortcite{li16} is intact. In fact, the introduction of the $\gamma$ parameter from Equation \ref{eq:catn} provides a convenient way to introduce semantic style transfer incrementally.

\section{Experiments}

\begin{figure}[!t]
\centering
\includegraphics[scale=1.45]{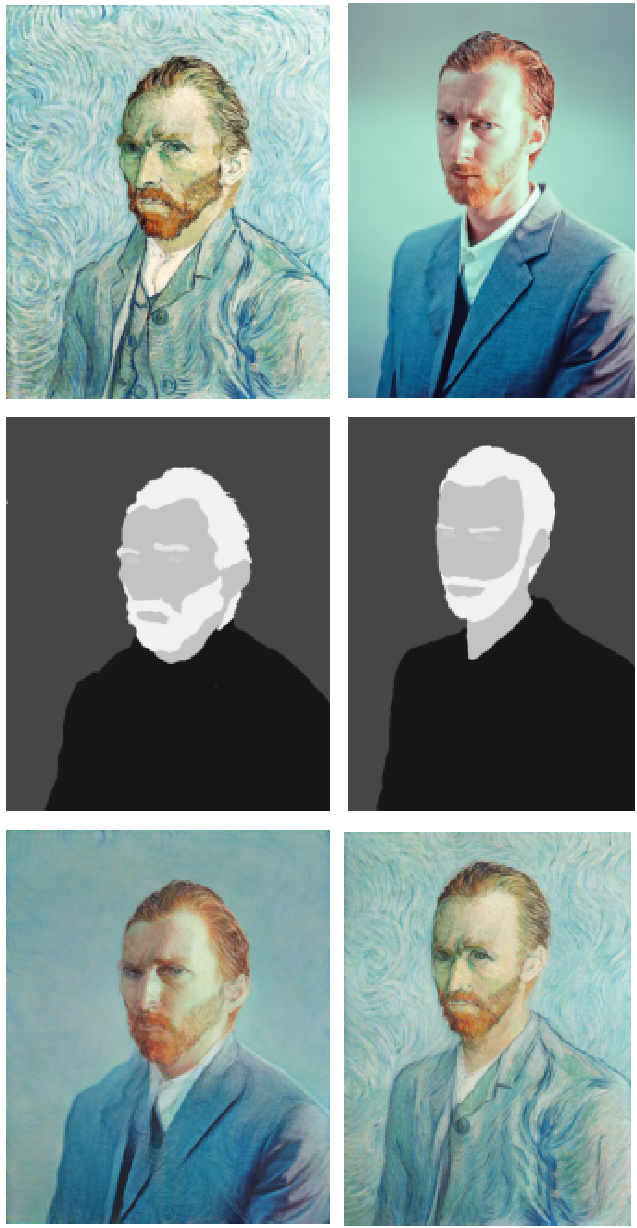}
\caption{Examples of semantic style transfer with Van Gogh painting. Annotations for nose and mouth are not required as the images are similar, however carefully annotating the eyeballs helps when generating photo-quality portraits. [Photo by Seth Johnson, concept by Kyle McDonald.]}
\label{fig:gogh}
\end{figure}

The following experiments were generated from VGG19 network using augmented layers \texttt{sem3\_1} and \texttt{sem4\_1}, with $3 \times 3$ patches and no additional rotated or scaled versions of the style images. The semantic maps used were manually edited as RGB images, thus channels are in the range \texttt{[0..255]}. The seed for the optimization was random, and rendering completed in multiple increasing resolutions---as usual for patch-based approaches~\cite{li16}. On a GTX970 with 4Gb of GPU RAM, rendering takes from 3 to 8 minutes depending on quality and resolution.

\subsection{Precise Control via Annotations}

Transferring style in faces is arguably the most challenging task to meet expectations---and particularly if the colors in the corresponding segments of the image are opposed. Typical results from our solution are shown in portraits from Figure \ref{fig:portraits}, which contains both success cases (top row) and sub-optimal results (bottom row). The input images were chosen once upfront and not curated to showcase representative results; the only iteration was in using the semantic map as a tool to improve the quality of the output.

In the portraits, the semantic map includes four main labels for background, clothing, skin and hair---with minor color variations for the eyes, mouth, nose and ears. (The semantic maps in this paper are shown as greyscale, but contain three channels.)


In practice, using semantic maps as annotations helps alleviate issues with patch- or gram-based style transfer. Often, repeating patches appear when setting style weight $\beta$ too high (Figure \ref{fig:portraits}, second row). When style weight is low, patterns are not transferred but lightly blended (Figure \ref{fig:portraits}, first row).  The semantics map prevents these issues by allowing the style weight to vary relative to the content without suffering from such artifacts; note in particular that the skin tones and background colors are transferred more faithfully.

\subsection{Parameter Ranges}

\begin{figure}[!t]
\centering
\includegraphics[scale=1.04]{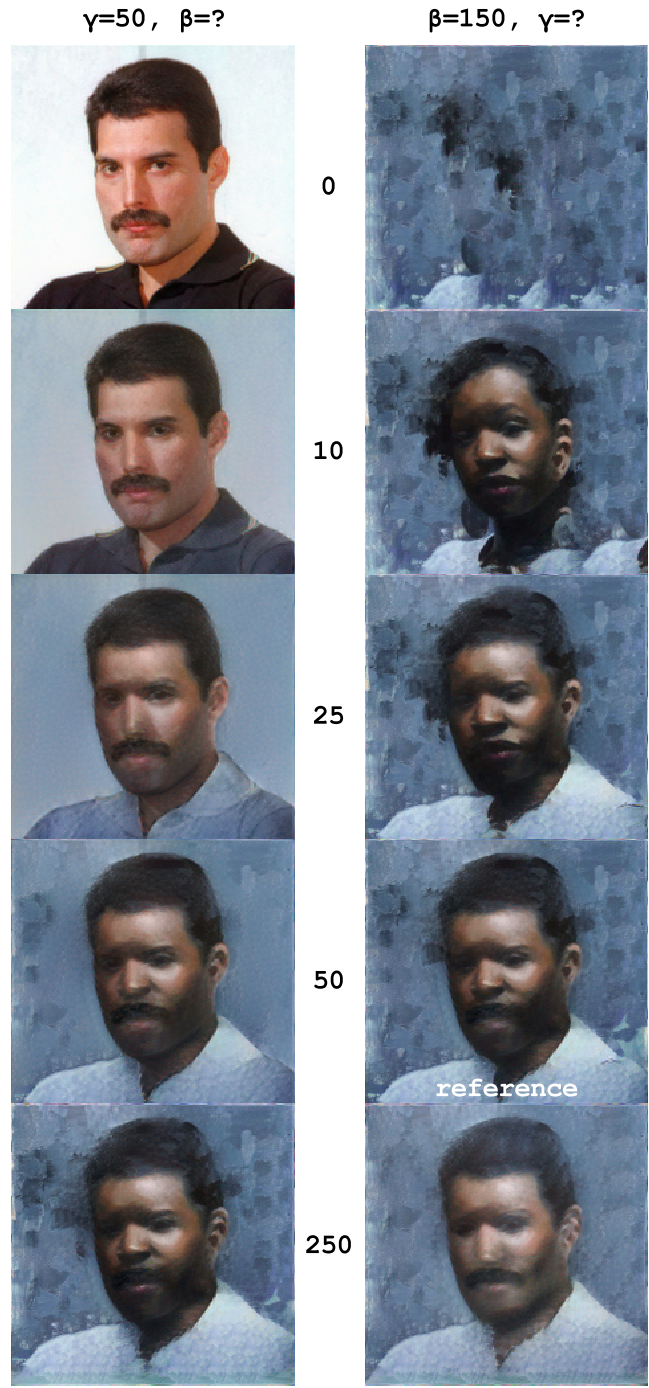}
\caption{Varying parameters for the style transfer. First column shows changes in style weight $\beta$: 0) content reconstruction, 10 to 50) artifact-free blends thanks to semantic constraint, 250) best style quality. Second column shows values of semantic weight $\gamma$: 0) style overpowers content without semantic constraint, 10) low semantic weight strengthens influence of style, 50) default value that equalizes channels, 250) high semantic weight lowers quality of style.}
\label{fig:params}
\end{figure}

Given a fixed weight for the content loss $\alpha = 10$, the style loss $\beta$ for images in this paper ranges from 25 to 250 depending on image pairs.  Figure~\ref{fig:params} shows a grid with visualizations of results as $\beta$ and $\gamma$ vary; we note the following:

\begin{itemize}
\item The quality and variety of the style degenerates as $\gamma$ increases too far, without noticeably improving the precision wrt. annotations.
\item As $\gamma$ decreases, the algorithm reverts to its semantically unaware version that ignores the annotations provided, but also indirectly causes an increase in style weight.
\item The default value of $\gamma$ is chosen to equalize the value range of the semantic channels $\bv{m}^l$ and convolution activations $\bv{x}^l$, in this case $\gamma = 50$.
\item Lowering $\gamma$ from its default allows style to be reused across semantic borders, which may be useful for certain applications if used carefully.
\end{itemize}

In general, with the recommended default value of $\gamma$, adjusting style weight $\beta$ now allows meaningful interpolation that does not degenerate into abstract patchworks.

\section{Analysis}

\begin{figure*}[!t]
\centering
\includegraphics[scale=1.8]{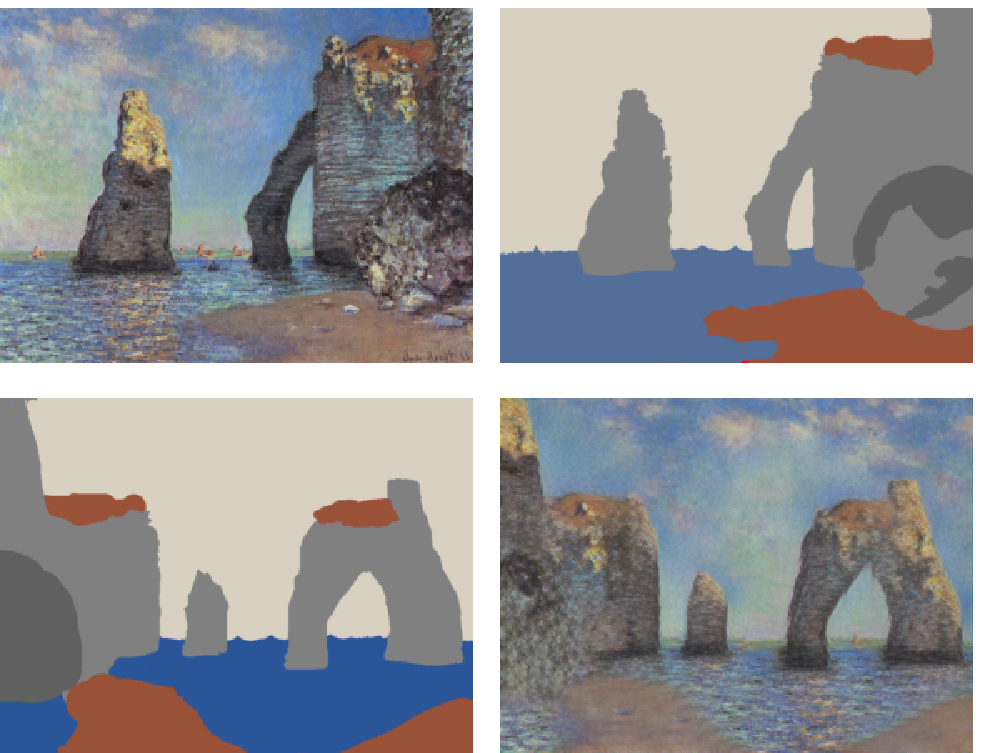}
\caption{Deep image analogy for a Monet painting based on a doodle; it's effectively semantic style transfer with no content loss.  This result was achieved in only eight attempts, showcasing the potential for the algorithm as an interactive tool.}
\label{fig:network}
\end{figure*}

Here we report observations from working with the algorithm, and provide our interpretations.

\subsubsection{Semantic Map Values} Since the semantic channels $\bv{m}^l$ are integrated into the same patch-based calculation, it affects how the normalized cross-correlation takes place. If the channel range is large, the values from convolution $\bv{x}^l$ will be scaled very differently depending on the location in the map.  This may be desired, but in most cases it seems sensible to make sure values in $\bv{m}^l$ have similar magnitude.

\subsubsection{Authored Representations} We noticed that when users are asked to annotate images, after a bit of experience with the system, they implicitly create ``semantic embeddings'' that compactly describe pixel classes. For example, the representation of a stubble would be a blend between hair and skin, jewelry is similar but not identical to clothing, etc. Such representations seem better suited to semantic style transfer than plain layer masks.

\subsubsection{Content Accuracy vs. Style Quality} When using semantic maps, only the style patches from the appropriate segment can be used for the target image. When the number of source patches is small, this causes repetitive patterns, as witnessed in parts of Figure \ref{fig:portraits}. This can be addressed by loosening the style constraint and lowering $\gamma$, at the cost of precision.

\subsubsection{Blending Segments} In the examples shown and others, the algorithm does a great job of painting the borders between image segments, often using appropriate styles from the corresponding image.  Smoothing the semantic map can help in some cases, however, crisp edges still generate surprising results.

\subsubsection{Weight Sensitivity} The algorithm is less fragile to adjustments in style weight; typically as the weight increases, the image degenerates and becomes a patchwork of the style content. The semantic map helps maintain the results more consistent for a wider range of the parameter space.

\subsubsection{Performance} Due to the additional channels in the model, our algorithm requires more memory as well as extra computation compared to its predecessor. When using only RGB images this is acceptable: around 1\% extra memory for all patches and all convolution output, approximately 5\% extra computation. However with pixels labeled using individual classes this grows quickly. This is a concern, although patch-based solutions in general would benefit from significant optimization that would apply here too.

\section{Conclusion}

Existing style transfer techniques perform well when colors and/or accuracy don't matter too much for the output image (painterly, abstract or sketch styles, glitch art) or when both image patterns are already similar---which obviously reduces the appeal and applicability of such algorithms.  In this paper, we resolved these issues by annotating input images with a semantic map, either manually authored or from pixel labeling algorithms. We introduced an augmented CNN architecture to leverage this information at runtime, while further tying advances in image segmentation to image synthesis. We showed that existing patch-based algorithms require minor adjustments and perform very well using this additional information.

The examples shown for style transfer show how this technique helps deal with completely opposite patterns/colors in corresponding image regions, and we analyzed how it helps users control the output of these algorithms better. Reducing the unpredictability of neural networks certainly is a step forward towards making them more useful as a tool to enhance creativity and productivity.

\section{Acknowledgments}

This work was possible thanks to the \textbf{nucl.ai Conference 2016}'s marketing budget that was repurposed for this research. As long as you visit us in Vienna on July 18-20, it's better this way! \url{http://nucl.ai}

Thanks to Petra Champandard-Pail, Roelof Pieters, Sander Dieleman, Jamie Blondin, Samim Winiger, Timmy Gilbert, Kyle McDonald, Mia Bergeron, Seth Johnson.

\bibliographystyle{iccc}
\bibliography{iccc}


\end{document}